%% file: root.tex
\documentclass[letterpaper, 10 pt, conference]{ieeeconf}  % Comment this line out if you need a4paper

\IEEEoverridecommandlockouts                              % This command is only needed if 
\usepackage{cite}
\usepackage{graphics} % for pdf, bitmapped graphics files
\usepackage{epsfig} % for postscript graphics files
\usepackage{times} % assumes new font selection scheme installed
\usepackage{amsmath} % assumes amsmath package installed
\usepackage{amssymb}  % assumes amsmath package installed
\usepackage{graphicx}
\usepackage[dvipsnames]{xcolor}
\usepackage{comment}   
\usepackage{caption}
\usepackage{subcaption}    
\usepackage{xcolor}
\usepackage{xfrac}
\usepackage{tabularray}
\usepackage{pgfplots}
\usepackage{tikz, tikz-3dplot}
\usetikzlibrary{math}
\usetikzlibrary{tikzmark}
\usepackage{xparse}
\usepackage{float}
\usepackage[linesnumbered,ruled]{algorithm2e}
\usepackage{url}
\usepackage{bm}
\usepackage{mathtools}

\pgfplotsset{compat=newest}

\SetCommentSty{mycommfont}

\DeclarePairedDelimiter{\paren}{(}{)}
\DeclarePairedDelimiter{\norm}{\|}{\|}

\NewDocumentCommand{\shoulder}{}{\textit{s}}
\NewDocumentCommand{\upperarm}{}{\textit{u}}
\NewDocumentCommand{\forearm}{}{\textit{f}}
\NewDocumentCommand{\elbow}{}{\textit{e}}
\NewDocumentCommand{\wrist}{}{\textit{w}}
\NewDocumentCommand{\vect}{m}{\bm{#1}}

\title{\LARGE \bf
An Avatar Robot Overlaid with the 3D Human Model of a Remote Operator}

\author{Ravi Tejwani$^{1}$, Chengyuan Ma$^{1}$, Paolo Bonato$^{2}$ and H. Harry Asada$^{3}$,$~\IEEEmembership{Fellow,~IEEE}$%
\thanks{$^{1}$Ravi Tejwani and Chengyuan Ma are with Department of Electrical Engineering and Computer Science (EECS), Massachusetts Institute of Technology,
Cambridge, MA 02142 USA
        {\tt\small \{tejwanir, macy404\}@mit.edu}}%
\thanks{$^{2}$Paolo Bonato is with Harvard Medical School Department of Physical Medicine and Rehabilitation at Spaulding Rehabilitation Hospital, 
Charlestown, MA 02129 USA
        {\tt\small pbonato@mgh.harvard.edu}}%
\thanks{$^{3}$H. Harry Asada is with Department of Mechanical Engineering, Massachusetts Institute of Technology,
Cambridge, MA 02142 USA
        {\tt\small asada@mit.edu}}%
}

\begin{document}

\maketitle
\thispagestyle{empty}
\pagestyle{empty}

%%%%%%%%%%%%%%%%%%%%%%%%%%%%%%%%%%%%%%%%%%%%%%%%%%%%%%%%%%%%%%%%%%%%%%%%%%%%%%%%
\begin{abstract}
Although telepresence assistive robots have made significant progress, they still lack the sense of realism and physical presence of the remote operator. This results in a lack of trust and adoption of such robots.
In this paper, we introduce an Avatar Robot System which is a mixed real/virtual robotic system that physically interacts with a person in proximity of the robot. The robot structure is overlaid with the 3D model of the remote caregiver and visualized through Augmented Reality (AR). In this way, the person receives haptic feedback as the robot touches him/her. 
We further present an Optimal Non-Iterative Alignment solver that solves for the optimally aligned pose of 3D Human model to the robot (shoulder to the wrist non-iteratively). The proposed alignment solver is stateless, achieves optimal alignment and faster than the baseline solvers (demonstrated in our evaluations).
We also propose an evaluation framework that quantifies the alignment quality of the solvers through multifaceted metrics. 
We show that our solver can consistently produce poses with similar or superior alignments as IK-based baselines without their potential drawbacks.
\end{abstract}

%%%%%%%%%%%%%%%%%%%%%%%%%%%%%%%%%%%%%%%%%%%%%%%%%%%%%%%%%%%%%%%%%%%%%%%%%%%%%%%%
\section{INTRODUCTION}

% Introduce the problem
In a situation when a person (such as an older adult) needs physical assistance, a care giver would be required to provide that support. In remote interactions, physical assistance should be delivered without degrading the quality of service. However, providing quality remote service is more challenging than in-person care. This could be realized by using a physical agent, i.e. a robot, being operated from a remote site. However, it might be uncomfortable or even terrifying to have such robot touch a care recipient's body. Rather, it is desirable that a remote caregiver appears to be in contact with the recipient while the robot is providing the physical assistance. 

% Current advancements made to address this problem
Advancements in telepresence assistive robots have enabled care givers to remotely monitor and assist patients during the rehabilitation process by observing and communicating with patients in real-time, even if they are not physically present in the same location. These robots have been used in a variety of rehabilitation services, including exercises, stretches, and massages ~\cite{helal2006tecarob, koceski2016evaluation, thacker2005physician}. They have also helped patients to improve their mobility and balance by providing feedback on their movements and posture ~\cite{bernardoni2019virtual, ballester2015accelerating, brutsch2011virtual}.

% How the current robots suck and there is a need for a new system
The research on visual representation of the remote therapist on to the robot has been limited in their ability to display the remote therapist in the form of 2D screens \cite{helal2006tecarob, koceski2016evaluation, agarwal2007roboconsultant} or using virtual and augmented reality to display the remote person \cite{jones2020vroom, neustaedter2016beam, erickson2020vrassistive}. Even the state of the art telepresence avatar robots \cite{lenz2023bimanual, luo2022towards, AvatarXPRIZE} simply project the avatar (3D human model) of the remote therapist on to the screens.
Hence, there is a fundamental disconnect in the research literature between the visual display of the remote care giver on the tele-operated robot and the actual remote physical therapy performed by the robot. 
In this paper, we aim to bridge this gap by conceptualizing the development of an Avatar Robot as a mixed real/virtual assistive tele-operated robot -- a physical robot that is overlaid with the 3D Human model of a remote care giver.
Therefore, the care receiver might feel as though he is in contact with the remote operator.

% The Avatar Robot system includes an end-to-end real-time pipeline that comprises trajectory control of the robot, cross-device pose synchronization, robot tracking, alignment solver(s), and online 3D projection overlay using an Augmented Reality (AR) device. 

% We further present an Optimal Non-Iterative Alignment solver \footnote{Code is available at \textcolor{blue}{\protect\url{https://github.com/Avatar-Robots/alignment/}}} that solves for the optimally aligned pose of 3D Human model to the robot (shoulder to the wrist). 
%
% Compared to the baselines adapted from IK solvers, our solver is specialized for alignment tasks, concise in form, fast in execution, stable and singularity-free. 

Our contributions include: 
\begin{enumerate}

    \item \textit{Avatar Robot System:} To the best of our knowledge, we propose the first general purpose end-to-end robotic system that provides the integration of the trajectory control of the robot, cross-device pose synchronization, robot tracking, alignment solver(s), and online 3D projection overlay using an Augmented Reality (AR) device; it achieves efficient and reactive real-time pose synchronization, alignment computation, and overlay projection.
    
    \item \textit{Alignment solver:} An optimal non-iterative alignment solver that solves for the optimally aligned pose of 3D Human model to the robot (shoulder to the wrist). We show how existing iterative IK solvers such as Jacobian DLS \cite{buss2004introduction} and FABRIK \cite{aristidou2011fabrik} can be adapted for this purpose. Compared to these baselines, our solver is intuitive, concise, deterministic, singularity-free, and superior in alignment metrics. We also provide theoretical guarantees of the solver;

    \item \textit{Empirical metrics and evaluation:} Set of evaluation metrics that qualitatively measures the quality of human-robot alignment from multiple perspectives including visual overlap, structural deviation, and model distortion. Our evaluation metrics could serve as a benchmark for evaluating future telepresence robots and their performance.
    
\end{enumerate}

% In this paper, we present a method for care receiver to perceive that the real caregiver is physically interacting with him/her while the robot is providing the physical assistance. 

\section{AVATAR ROBOT SYSTEM}\label{sec:preliminary}

\begin{figure}[t]
    \centering
    \begin{subfigure}[b]{.4\linewidth}
        \captionsetup{justification=centering}
        \includegraphics[width=0.87\linewidth]{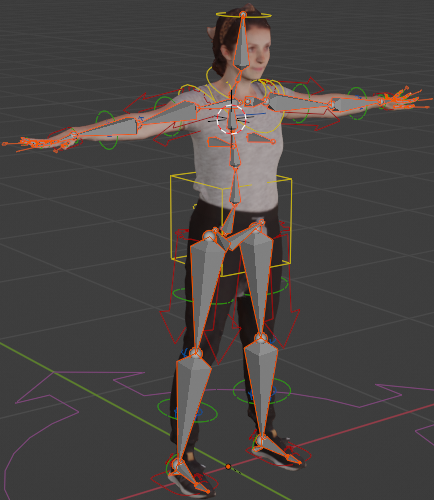}
        \caption{3D Scan of Physical Therapist}\label{fig:3d-scan-therapist}
    \end{subfigure}
    \setcounter{subfigure}{2}
    \begin{subfigure}[b]{.3\linewidth}
        \captionsetup{justification=centering}
        \includegraphics[width=\linewidth]{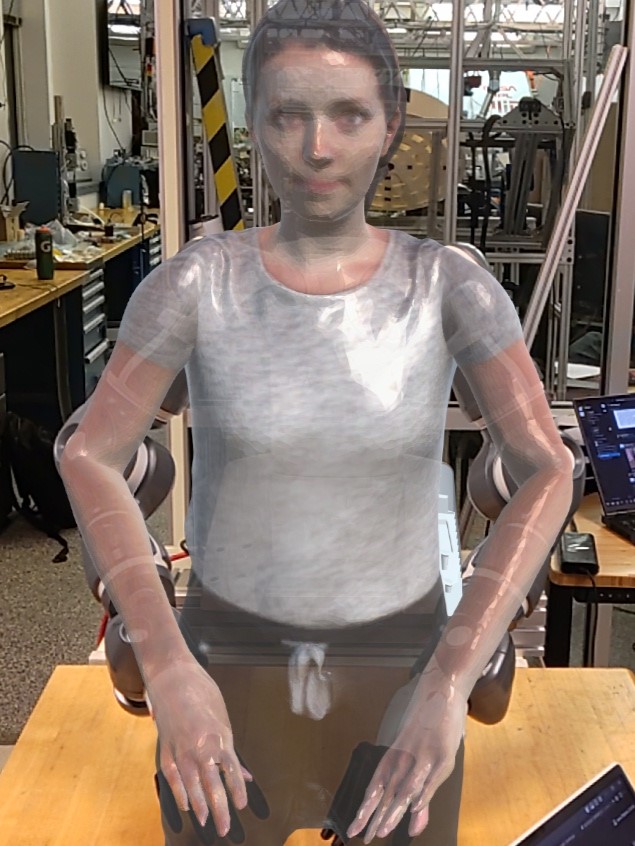}
        \caption{Aligned Human model over robot}\label{fig:teaser-image}
    \end{subfigure}
    \setcounter{subfigure}{1}
    \begin{subfigure}[b]{.3\linewidth}
        \captionsetup{justification=centering}
        \includegraphics[width=\linewidth]{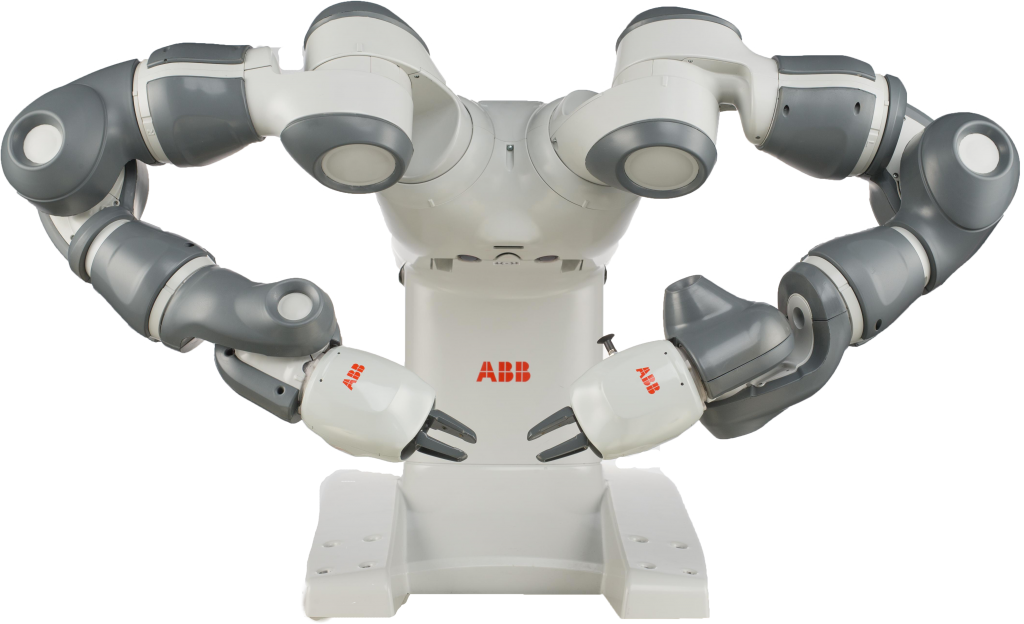}
        \caption{Collaborative robot (YuMi)}\label{fig:yumi-robot}
    \end{subfigure}
    
    \caption{Aligning Human model with Collaborative robot}
    \label{fig:problem}
\end{figure}

An Avatar Robot is a mixed real/virtual robotic system that physically interacts with a person in proximity of the robot. As shown in Fig. \ref{fig:problem}, a remotely operated robot and the 3D model of a remote operator are incorporated into our system. The robot structure is overlaid with the 3D model of the remote caregiver using our solvers and visualized through Augmented Reality (AR). In this way, the person receives haptic feedback as the robot is in contact with him/her. Simultaneously, the person would visually observe the caregiver’s presence through 3D rendering. Thus, care receivers will have a highly immersive experience when haptic stimuli and visual observations are synchronized.

\textbf{Problem} Given a 3D model of a human, obtained from the 3D scan of the physical therapist (Fig. \ref{fig:3d-scan-therapist}), and a dual arm collaborative YuMi robot (Fig. \ref{fig:yumi-robot}) which can provide physical assistance to the person. 
Our objective is to optimally align and overlay them across their joints, axes and rotations as the Avatar Robot interacts with the person.(Fig. \ref{fig:teaser-image}).

% \section{Preliminaries}\label{sec:preliminary}

\textbf{Human Model} We prepare the 3D human model from a photogrammetry scan of a Physical Therapist from Spaulding Rehabilitation Center, Charlestown, MA. The therapist was in T-pose at the time of scan, and we manually annotated the joints of the model so the model can be posed dynamically. In this research, we are interested in dynamically computing the pose of the model's arms to create optimal alignment with the robot arms, so joints of particular concern are the shoulder, elbow, and wrist joints on both sides. %Treating the elbows as hinge joints, we also marked their axes of rotation (later referred to as ``elbow axis''). 

\textbf{Structural Correspondence between the Human Model and the Robot}
Since our robot has anthropomorphic arms, with clearly identifiable ``upper arm'' and ``forearm'' segments, our task is to align the human model's shoulder, elbow, and wrist joints to the corresponding \emph{reference positions} on the robot, which we choose to be the center of axis 2, 3, 6 of the robot respectively. A side-to-side comparison between the arms of our human model and the robot as well as the reference positions that we chose is shown in Fig. \ref{fig:yumi-vs-human-arms}. 

As a preliminary step of alignment, we also compute the position and rotation of the entire human model such that the segment connecting its shoulder joints overlaps and share the same midpoint with the segment connecting the shoulder reference positions on the robot. This step ensures the body /shoulder of the human model is aligned with the body of the robot at large and allows us to compute the best-aligning pose of two arms separately.

\begin{figure}[b]
    \centering
    \includegraphics[trim=0 3cm 11.7cm 0, clip]{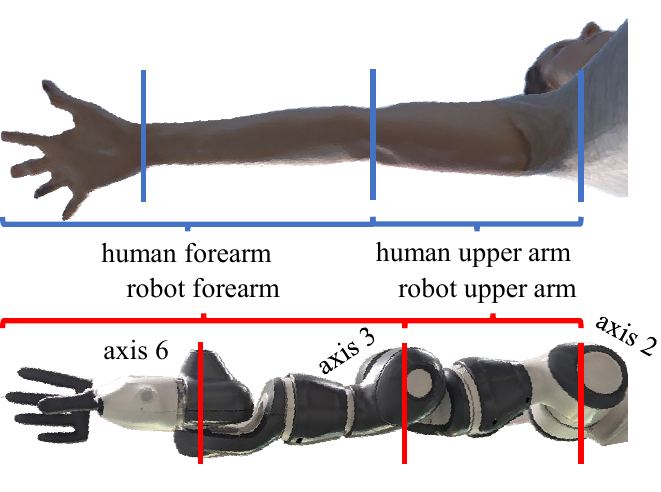}
    \caption{Side-by-side comparison of our robot arm and the arm of our human model. We set the center of axis 2, 3, and 6 of our robot as the shoulder, elbow, and wrist reference points for the human model. The difference in proportions of arm segments between the human model and the robot makes scaling necessary in order to achieve good alignment.}
    \label{fig:yumi-vs-human-arms}

\iffalse
    \smallskip
    \begin{minipage}{\linewidth}
        \linespread{1}
        \footnotesize(YuMi's axis 3 is slightly off the arm axis shown above. Hence setting it as elbow reference directly is problematic. An alternative is the intersection of center lines of two arm segments, but that proved to cause excessive scaling. We empirically set a weighted average of the two as the true elbow reference.)
    \end{minipage}
\fi
\end{figure}

\begin{figure*}
    \centering
    \includegraphics{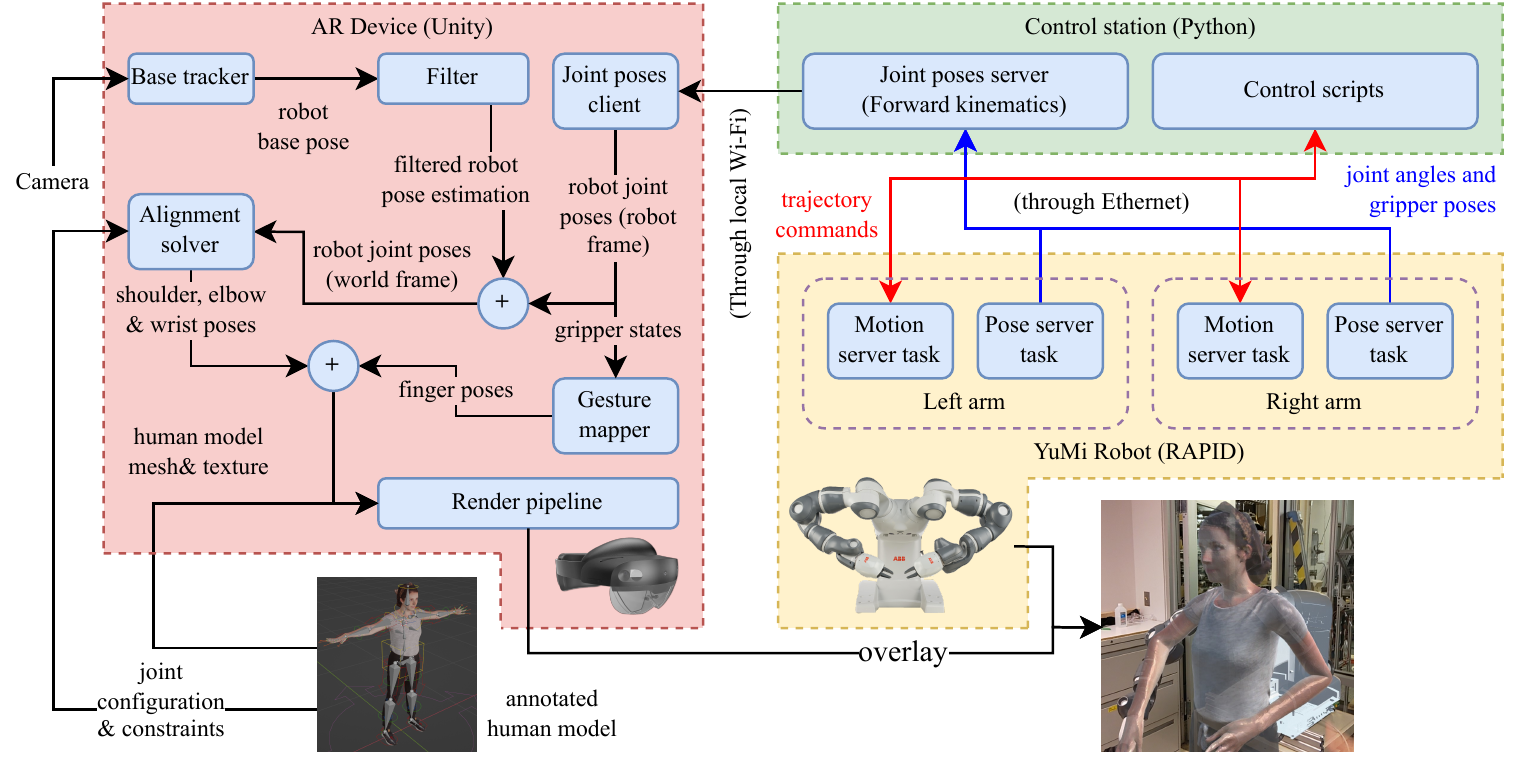}
    \caption{The pipeline of our Avatar Robot System. A motion server and a pose server run in parallel on both arms of the robot. 
    The motion server executes commands from scripts on the control station. 
    The control station also runs a relay that aggregates joint angles from both arms and stream dual-arm joint poses to the AR device over WLAN. 
    The AR device converts local-frame joint positions to world-frame by them combining with the pose estimate of the robot from its cameras. 
    An alignment solver then solves for the human model pose;
    A gesture mapper maps gripper positions to hand gestures of the human model. 
    With the arm poses and gesture, the renderer renders the human model accurately over the robot.
    Joint poses are synchronized between robot, AR device and the control station at 100Hz, while the rendering is at 60Hz.
}
\label{fig:system-diagram}
\end{figure*}

\textbf{Defining Input and Output for the Solvers} The input to an alignment solver consists of an joint positions of the human model in its default (unaligned) configuration and the reference joint positions of the robot. The output will be an \emph{arm pose} for each side, which include the rotations of joints as well as stretch factor of forearm and upper arm of the human model. Allowing stretching parts of the human model risks making the output pose un-anthropormorphic, but it is necessary for better alignment as the proportions of robot and human arms are inevitably different (see Fig. \ref{fig:yumi-vs-human-arms}). We assume that some extent of scaling is acceptable. We will revisit incorporating scaling and other constraints in the description of the solvers.

\textbf{Pipeline} The Avatar Robot System integrates the alignment solvers into an end-to-end real-time workflow that comprises trajectory control of the robot, cross-device pose synchronization, robot tracking, alignment solving, and overlay projection. The pipeline of robot System is presented in Fig. \ref{fig:system-diagram}. It consists of three components: the robot, the control station, and the augmented reality (AR) device. 

The robot, a dual-arm collaborative robot \cite{yumi}, runs a pose server and a generic motion server in parallel on each arm, and communicates with the control station through Ethernet. 
The control station stores control scripts that issues specific trajectory commands to the robot to perform various tasks, e.g., emulating a rehabilitation therapy. It also collects joint angles and gripper orientations from both arms at real-time and hosts a relay server to streatm this information through the local wireless network.
We use HoloLens 2 \cite{hololens} as the AR device. The pose of the robot base is tracked using the camera mounted on the AR device with a computer vision model. 
The raw pose estimation is further passed to a low-pass filter to reduce noises and stuttering. The filtered pose is combined with the robot-frame joint coordinates retrieved from the control station real-time to compute the world-frame joint positions. Our system then computes the shoulder, elbow, and wrist reference positions for the human model and aligns its body with the robot. The reference positions are then passed to a alignment solver, whose pose output is then fed to the rendering pipeline which renders the human model and projects it over the robot in front of the patient.  

% As for implementation, we used the YuMi dual-arm collaboration robot and HoloLens 2\rev{cite it} as the testing platform for our system\cite{yumi}. The AR-device part of our system is written with Unity and Vuforia framework, and the control station part is written in Python \cite{CITE EVERYTHING HERE}. 

For completeness, we also designed a gripper mapper module that maps the state of robot gripper into the gesture of the model. Our gripper is based on a 1DoF linear gripper with fingers customized to resemble thumb and fingers, so we map the position of the gripper linearly to the angle between the thumb and index finger of the model.
In practice, this system can align a human model pose with a robot in real-time with minimal latency and high accuracy.

\iffalse
\begin{figure*}[ht]
    \centering
    \includegraphics[trim=0 3cm 5cm 0, clip]{figures/yumi-vs-human-arms.pdf}
    \caption{Side-by-side comparison of our robot arm and the arm of a human model. We designate the center of axis 2 and 6 as the shoulder and wrist target positions for the human model. The difference in dimensions between arm segments of the human model and the robot makes stretching necessary our solver in order to achieve good alignment. The elbow reference is determined as shown on the right: alignment would be suboptimal if we align the model's elbow directly to axis 3 (as shown in red); If we align elbow to the intersection of center lines of the two robot arm linkages (shown in green), human arms will be excessively stretched. As a compromise, we set the elbow reference of the human model to $0.8\,\text{(intersection)}+0.2\,\text{(axis center)}$, shown as the blue dot above. }
    \label{fig:yumi-vs-human-arms}
\end{figure*}
\fi

\section{SOLVERS}
In this section, we first explore how existing iterative Inverse Kinematics (IK) solvers can be adapted to perform the aforementioned alignment task. Specifically, we show adaptation of two IK solvers: Jacobian Damped Least Square \cite{buss2004introduction}, commonly used in robotics, and FABRIK \cite{aristidou2011fabrik}, commonly used in humanoid animation.
These solvers serve as our baselines. We then introduce our optimal non-iterative alignment (ONIA) solver.

% These baselines might be valuable because they are more generic and allow constraints on the generated poses of the human model, which may be helpful if we want to keep the human model's pose more natural and anthropomorphic at the cost of visual alignment. On the other hand, iterative solvers risk instability, require careful tuning of hyper parameters and are known to get stuck in singularities.

\subsection{Inverse Kinematic Solvers}
%Inverse Kinematics solvers require a clear specification of the kinematic chain. We model the shoulder joint of a human model as a ball-and-socket joint whose rotation can be decomposed into a swing and a twist\cite{baerlocher2001parametrization}. 
%The elbow joint is seen as a composition of a hinge joint followed by a small range of twist around the forearm. This extra twist allows the skin around the elbow to react more realistically to large twist of the wrist. 

\subsubsection{Jacobian (adapted)}
\NewDocumentCommand{\trans}{}{^\mathsf{T}}
We first adapted the Jacobian method with damped least square (DLS) for the alignment task. Given a parameterization $\vect \theta$ and the current end effector position $\vect p$, the effect of perturbing $\vect \theta$ on $\vect p$ can be linearly approximated by $\vect{\Delta p} = J\vect{\Delta \theta}$ in $\vect p$'s neighborhood, where $J$ is the Jacobian. Provided with a desired $\vect{\Delta p}$, Jacobian with DLS computes $\vect{\Delta\theta}$ as follows,
\begin{equation}
    \vect{\Delta\theta} = J\trans \paren*{JJ\trans + \lambda^2I}^{-1}\vect{\Delta p}.
\end{equation}
where the hyper-parameter $\lambda$ is the damping constant that must be chosen with care to keep the solver both responsive and stable.

We model the human arm as the shown in Fig. \ref{fig:kinematics}. In particular, we parameterize the swing of the upper arm around the shoulder as an exponential map \cite{baerlocher2001parametrization, grassia1998practical}. To incorporate the reference position of the elbow in Jacobian-based algorithms, we start by noting that the Jacobian method is equivalent to gradient descent to the objective %\vspace{-1em}
\begin{equation} 
    \frac12{\norm*{\vect{p} - \vect{x}}^2}%\norm*{\vect{p}} - \vect{x}}^2 \overbrace{+\,\frac{\lambda^2}{2} \norm*{\vect{\Delta \theta}}^2}^{\smash{\text{for DLS}}}
\end{equation}
where $\vect{x}$ is the target position. For a joint, $\vect{p}_\elbow$, we want to be close to its reference, $\vect{x}_\elbow$, we add the term $\frac{1}{2}w_\elbow^2\norm*{\vect{p}_\elbow - \vect{x}_\elbow}^2$ to the objective, 
\begin{align}
    \frac12\norm*{\vect{p} - \vect{x}}^2 + \frac{1}{2}w_\elbow^2\norm*{\vect{p}_\elbow - \vect{x}_\elbow}^2 = \frac{1}{2}\norm*{\begin{bmatrix}
        \vect{p}\\w_\elbow\vect{p}_\elbow
    \end{bmatrix} - \begin{bmatrix}
        \vect{x}\\w_\elbow\vect{x}_\elbow
    \end{bmatrix}}^2
\end{align}
where $w_\elbow$ is a small constant so $\frac{1}{2}w_\elbow^2\norm*{\vect{p}_\elbow - \vect{x}_\elbow}^2$ becomes a secondary objective. %As shown above, the new objective can be rewritten to have the same form as the old one, and hence requires no change to the rest of the algorithm. %From another perspective, our approach can be seen as treating the joint with reference as another end effector and weighing end effectors differently, hence the name ``Weighted Multi-end-effector Jacobian.'' 
We find $w_\elbow = 0.1, \lambda = 0.2$ to work best. %Another problem we noticed was that initializing the human model in its default configuration (T-pose) often traps the solver in singularities, which we solve by bending the elbow by 90 degrees initially.
\subsubsection{FABRIK (adapted)}
Forward And Backward Reaching Inverse Kinematics (FABRIK) is a heuristic-based IK solver that received wide use in humanoid animation due to its simplicity and its natural poses \cite{aristidou2011fabrik}. Every iteration of FABRIK consists of a forward and backward reach that traverse the kinematic chain in different directions. It repositions a joint on the line connecting its current position and the preceding joint's updated position.

Our adaptation presented here enhances plain FABRIK by allowing every joints to have a reference positions. It produces a pose that minimizes the maximum deviation of any joint to its reference\footnote{We make the remark that a joint with a reference is different from an end effector with a target, because an end effector is assumed to be always able to reach its target in FABRIK but it may very well be infeasible for every joint to be at its reference position, and FABRIK is unstable when confronted with infeasbility. Hence, plain FABRIK's ability to handle multiple end effector does not help in our scenario.}. Under our adaptation, a joint with a reference position is given a deviation tolerance $\epsilon$. In every iteration, the joint is re-positioned not on the segment connecting its child and itself, but on the segment connecting the child and
the point closest to the joint \emph{within a radius-$\epsilon$ sphere around the reference position}. 
This heuristic is intuitive, local, minimal, easy to implement, and works well empirically. %It allows us to see the end effector as having the target as the reference position with a near-zero tolerance. 
The choice of $\epsilon$ is critical: if too large, the constraint is too loose to be useful; if too small, the solver may be unstable due to infeasibility. We determine the optimal $\epsilon$ by binary search.
% \rev{, maintaining the best feasible solution in the process}. 
% In practice, we empirically choose our prior of $\epsilon_1 = 30\operatorname{cm}$, $\Delta\epsilon = 0.2\operatorname{cm}$, $n_\textit{init} = 8$ and $n_\textit{refine} = 5$.

Both Jacobian and FABRIK allow incorporating joint rotational constraints into the solution. Hence, we explored adding basic human joint constraints into both baselines: We allow the upper arm to swing for at most $85^{\circ}$ in all directions and twist for $\pm 75^{\circ}$ relative to the T-pose. The hinge joint permits rotation from 0 to $150^{\circ}$ inward, with the T-pose as the zero reference. These constraints are more permissive than anatomical ground truth \cite{baerlocher2001parametrization} so as not to over-constrain our model, as alignment is also a tight constraint on its own.  We allow a stretch factor from 0.8 to 1.3 for both upper and forearm for reasons discussed in section \ref{sec:preliminary}.

\subsection{Optimal Non-Iterative Alignment Solver}

%The non-iterative solver, introduced below, achieves best alignment by assuming these constraints are always met. We expect hence the non-iterative solver and the baseline to deviate when the robot pose is not anthropomorphic on its own. In that case, whether non-aligning but natural-looking pose from the baseline or the best-aligning pose from the non-iterative solver would create better realism is a subjective judgement that could be the subject of future user studies.

%
% Their complexity allows them to consider joint constraints, which may be helpful to produce natural-looking poses with inferior alignment in case the robot arm pose is not anthropomorphic.
%
% Still, in most cases, the robot arm poses should be anthropomorphic already, the benefit of complex IK solvers becomes negligible, and we should prioritize alignment. 
%

It is helpful when the input robot arm pose is not anthropomorphic to use IK solvers that are generic and consider joint constraints. However, when the robot arm poses are anthropomorphic, the benefits of these baseline IK solvers fails to outweigh their limitations: they are prone to singularities, and require careful tuning of its hyper parameters to be stable. 
We introduce an Optimal, Non-Iterative, Alignment solver (ONIA) that solves for the optimally aligned pose from shoulder to the wrist non-iteratively. 
ONIA is stateless, achieves optimal alignment and faster than the baseline solvers (demonstrated in the evaluation section). We describe ONIA's design below.

% We show the steps of ONIA below, with nomenclature defined in Table \ref{tbl:nomenclature}. 
\NewDocumentCommand{\Fromto}{m m}{\ensuremath{{\mathcal{R}}_{#1}^{#2}}}
\NewDocumentCommand{\Proj}{m m}{\ensuremath{\paren*{#1}_{\perp\, #2}}}

\paragraph*{Nomenclature} We begin by defining our notations. 
Let $\Fromto{\vect x}{\vect y}$ denote the rotation that rotates $\vect x$ to $\vect y$ around $\vect x \times \vect y$. In other words, $\Fromto{\vect x}{\vect y}$ is the rotation such that
\begin{equation}
    \Fromto{\vect x}{\vect y} \vect x = \vect y,\quad \Fromto{\vect x}{\vect y}(\vect x \times \vect y) = \vect x \times \vect y.
\end{equation}
In addition, let
\begin{equation}
    \Proj{\vect x}{\vect y} = \vect x - \frac{\vect x \cdot \vect y}{\norm{\vect y}^2}\vect y
\end{equation}
denote the projection of $\vect x$ onto the plane normal to $\vect y$.

As outlined in section \ref{sec:preliminary}. The input to ONIA has two parts: the joint positions of an arm of the human model (in its default unaligned pose), and the corresponding reference positions of the robot. For clarity, through out this section, we'll use the superscripts $h$ to denote joints from the human model and $r$ to denote references positions of the robot. We will also use subscripts $\shoulder$, $\elbow$, $\wrist$, $\upperarm$, $\forearm$ for houlder, elbow, wrist, upper arm, and forearm. All joints and rotations below are of the human model unless explicitly specified otherwise. Fig. \ref{fig:kinematics} visualizes the symbols we used below and presents the kinematic chain of the human model which ONIA works on.

ONIA aligns the given arm of the human model to an arm of the robot from shoulder to wrist through the following steps:

\begin{figure}[t]
    \centering
    \input{ref_latex/kinematics.tex} 
    \caption{The kinematics model. We model the shoulder joint of a human model as a ball-and-socket joint which can be decomposed into a swing and a twist\cite{baerlocher2001parametrization}. The elbow joint is composed of a hinge followed by a small of twist around the forearm. The shoulder of the human model and the robot were assumed to be aligned initially (see section \ref{sec:preliminary}). The notations are used by the description of our non-iterative solver below.}
    \label{fig:kinematics}
\end{figure}
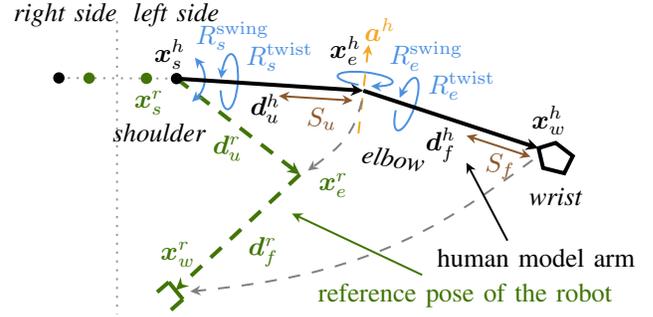

\iffalse
\paragraph*{Preliminary -- aligning the shoulder} Recall that we have aligned the shoulder of the human model to that of the robot, such that 
\begin{align}
    \vect x_\shoulder^{h, \text{left}} - \vect x_\shoulder^{h, \text{right}} 
    &\parallel \vect x_\shoulder^{r, \text{left}} - \vect x_\shoulder^{r, \text{right}},\ \text{and}\\ 
    \frac12 \paren*{\vect x_\shoulder^{h, \text{left}} + \vect x_\shoulder^{h, \text{right}}} &= \frac12
    \paren*{\vect x_\shoulder^{r, \text{left}} + \vect x_\shoulder^{r, \text{right}}}.
\end{align}
Because the human model's shoulder width differs from the robot's shoulder width, the human model's and robot's shoulder joints cannot be perfectly aligned at the same time.
\fi

\paragraph{Computing arm axes} ONIA begins by computing the \emph{arm axes} of both the human model and the robot, which are vector representations of the upper and forearm segments. Let $\vect{x}_\shoulder^h, \vect{x}_\elbow^h$ and $\vect{x}_\wrist^h$ be the positions of the human model's shoulder joint, and let $\vect{x}_\elbow^r$ and $\vect{x}_\wrist^r$ be the reference elbow and wrist positions of the robot. The elbow axes are computed as
\begin{align}
    \vect{d}_\upperarm^h &= \vect{x}_\elbow^h - \vect{x}_\shoulder^h,\qquad \vect{d}_\forearm^h = \vect{x}_\wrist^h - \vect{x}_\elbow^h,\\
    \vect{d}_\upperarm^r &= \vect{x}_\elbow^r - \vect{x}_\shoulder^h,\qquad 
    \vect{d}_\forearm^r = \vect{x}_\wrist^r - \vect{x}_\elbow^r.
\end{align}
Note $\vect{d}_\upperarm^r$ starts at $\vect{x}_\shoulder^h$ instead of $\vect{x}_\shoulder^r$ (see Fig. \ref{fig:kinematics}). Because the human model's shoulder width differs from the robot's shoulder width. The human model's and robot's shoulder joints cannot be perfectly aligned at the same time. Hence, we use $\vect{x}_\shoulder^h$ as the basis for guiding rotation of the upper arm.

% because its purpose is to guide ONIA in rotating the upper arm around $\vect{x}_\shoulder^h$, the shoulder of the human model.
\paragraph{Aligning direction of the upper arm} The direction of the upper arm is aligned with the reference axis $\vect d_\upperarm^r$ by swinging it around the shoulder,
\begin{align}
    R_\shoulder^{\text{swing}} = \Fromto{\vect{d}_\upperarm^r}{\vect{d}_\upperarm^h}. \label{eq:shoulder-swing}
\end{align}
%After applying $R_\shoulder^{\text{swing}}$, the upper arm axis of the human model should overlap with the reference upper arm axis $\vect d_\upperarm^r$. 

\paragraph{Aligning scale of the upper arm} We further stretch(shrink) the upper arm segment so that its elbow joint reaches the reference position $\vect x_\elbow^r$. This is done by computing the scaling factor $S_\upperarm$, 
\begin{equation}
    S_\upperarm = \frac{\|\vect{d}_\upperarm^r\|}{\left\| \vect{d}_\upperarm^h\right\|}, \label{eq:upperarm-scale}
\end{equation}
such that
\begin{equation}
 \vect{d}_\upperarm^r = R_\shoulder^{\text{swing}}(S_\upperarm\vect{d}_\upperarm^h).
\end{equation}

\paragraph{Aligning the elbow axis} Further we twist the the upper arm so that its \emph{elbow axis} aligns with the \emph{reference elbow axis} of the robot. 
The elbow axis of a human model's arm is the axis around which the elbow rotates, if treated as a hinge joint. We annotate the elbow axis when prepossessing the model and denote it as $\vect a^h$. The reference elbow axis of the robot is the cross product of its reference arm axes,
\begin{align}
    \vect{a}^r = \vect{d}_\upperarm^r \times \vect{d}_\forearm^r.
\end{align}
Consequently, the twist component of the human arm's shoulder joint is
\begin{align}
    R_\shoulder^{\text{twist}} &= \Fromto{\vect{a}_{\perp}^h}{\vect{a}^r},\label{eq:align-elbow-axis}\\
    \text{where}\ \vect{a}_{\perp}^h &= \Proj{R_\shoulder^{\text{swing}}\vect{a}^h}{\vect{d}_\upperarm^r}.
\end{align}
Essentially, we rotate the swung upper arm (and hence its elbow axis) around itself so the projection of the elbow axis onto the plane normal to the axis of rotation, $\vect{d}_\upperarm^r$, coincides with the reference elbow axis $\vect{a}^r$. This minimizes the angle between the human model's elbow axis and the reference elbow axis: in general, given vectors $\vect u, \vect v, \vect k \in \mathbb{R}^3$, where $\vect v$ can rotate around $\vect k$. The angle between $\vect v$ and $\vect u$ is minimized when $\Proj{\vect v}{\vect k}$ and $\Proj{\vect u}{\vect k}$ meet.

The overall rotation of the shoulder joint is the composition of the swing followed by the twist,
\begin{equation}
    R_\shoulder = R_\shoulder^{\text{twist}}R_\shoulder^{\text{swing}}. 
\end{equation}
We are now done with the shoulder joint and the upper arm and now move on to the elbow joint and the forearm. 

\paragraph{Aligning the direction of the forearm} Similar to \eqref{eq:shoulder-swing}, we align the forearm --- now $R_\shoulder\vect{d}_\forearm^h$ after rotating the shoulder --- by swinging it around the elbow so that it lines up with the reference forearm axis,
\begin{equation}
    R_\elbow^{\text{swing}} = \Fromto{R_\shoulder \vect{d}_\forearm^h}{ \vect{d}_\forearm^r}  R_\shoulder.\label{eq:elbow-swing}
\end{equation}
%Note that we are not rotating the forearm around the elbow axis of the human model. Instead, after the applying $R_\elbow^{\text{swing}}$, the forearm will effectively have rotated around the reference arm axis $\vect{a}'$. This distinction is negligible, as we already aligned the model's elbow axis to its reference in \eqref{eq:align-elbow-axis}.

\paragraph{Aligning the scale of the forearm} We further stretch (shrink) the forearm, similar to \eqref{eq:upperarm-scale}, so that wrist joint of the human model reaches the reference wrist position. The scaling factor of the human model's forearm is computed as
\begin{equation}
    S_\forearm = \frac{\|\vect{d}_\forearm^r\|}{\|\vect{d}_\forearm^h\|}.\label{eq:forearm-scale}
\end{equation}

\paragraph{Aligning the wrist} We set the rotation of the wrist $R_\wrist$ to be the same as the end effector rotation of the robot. 

\paragraph{Reacting to wrist twist} Finally, we twist the forearm slightly around itself proportional to the twist of the wrist joint. This leads to more realistic rendering by enabling the forearm skin near the elbow to move in response to the wrist twist. We develop the notion of \emph{wrist twist} through the following insight: Suppose in the default configuration of the human model, there is some ``wrist vector'' $\vect w$ that is roughly perpendicular to the palm. Then, after the wrist joint rotates by $R_\wrist$, the twist component of $R_\wrist$ is the rotation around the forearm axis that minimizes the angle between $\vect w$ and $R_\wrist \vect w$. Similar to \eqref{eq:align-elbow-axis}, this is the rotation that brings the projection of the two vectors together. With this insight, we formulate the proportional twist of the elbow joint as follows,
\begin{align}
    R_\elbow^{\text{twist}} &= \paren*{\Fromto{\vect{w}_{\perp}^h}{\vect{w}_{\perp}^r}}^{\alpha_\elbow}\\
    \text{where}\ \vect{w}_{\perp}^h &= \Proj{R_\elbow^{\text{swing}}\vect{w}}{\vect{d}_\forearm^r}, \vect{w}_{\perp}^r = \Proj{R_\wrist\vect{w}}{\vect{d}_\forearm^r}.
\end{align}
%Note that we use $\vect{w}_{\perp}^h = \Proj{R_\elbow^{\text{swing}}\vect{w}}{\vect{d}_\forearm^r}$ rather than $\vect{w}_{\perp}^h = \Proj{\vect{w}}{\vect{d}_\forearm^r}$ because we have already rotated the forearm (and hence the wrist normal) by $R_\elbow^{\text{swing}}$. 
We set $\alpha_e = 0.4$ empirically. Together, the overall rotation of the elbow joint is
\begin{equation}
    R_\elbow = R_\elbow^{\text{twist}}R_\elbow^{\text{swing}},
\end{equation}
This concludes ONIA. Note that in the above procedure, ONIA starts with the default configuration of the human arm and is not dependent on the current poses. Being non-iterative makes the algorithm easy to analyze and ensures a one-to-one, deterministic mapping from the robot poses to the human model poses. 

ONIA produces optimal alignment in two regards. First, the swinging (\eqref{eq:shoulder-swing} and \eqref{eq:elbow-swing}) and scaling (\eqref{eq:upperarm-scale} and \eqref{eq:forearm-scale}) ensures that the elbow and wrist joints of the human model reaches their reference positions on the robot \emph{exactly}. In addition, the twist in \eqref{eq:align-elbow-axis} \emph{minimizes} the angle between the human model's elbow axis and the robot's elbow axis.  

\begin{figure}
    \centering
    \includegraphics[width=\linewidth]{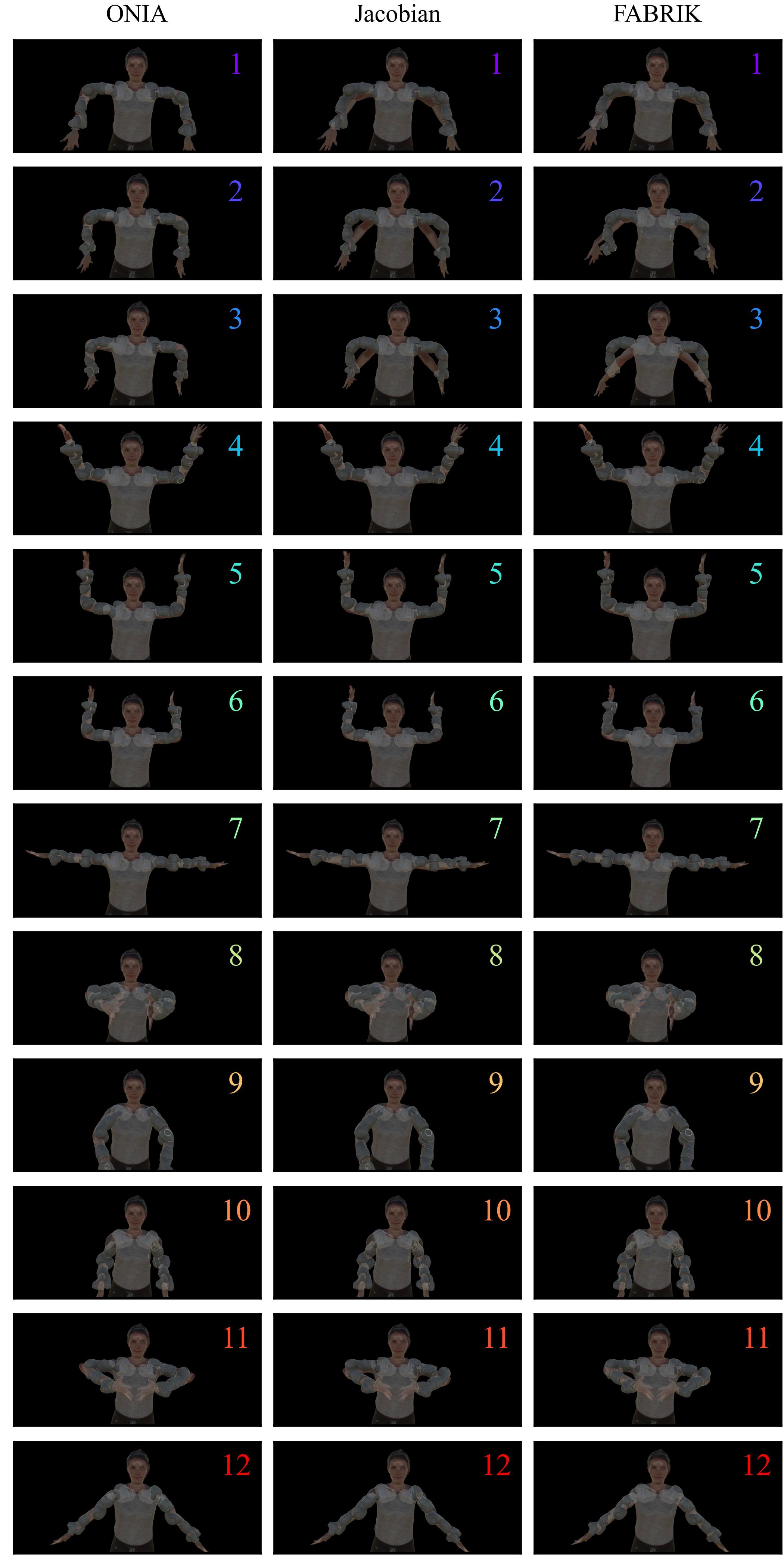}
    \iffalse
    \begin{subfigure}{\linewidth}
        \includegraphics{figures/poses1-9.pdf} 
        \caption{Poses 1 to 9 (ONIA)}
    \end{subfigure}
    \begin{subfigure}{\linewidth}
        \includegraphics{figures/poses10-12.pdf} 
        \caption{Poses 10 to 12 by each solver}
    \end{subfigure}
    \fi
    \caption{We evaluated the performance of the solvers on a set of 12 poses. We chose these poses because they are representative of the space of possible arm poses. Pose 1 to 3 all cover ``hands down'' but differ in forearm orientations. Pose 4 to 6 likewise covers ``hands up''. Pose 8 to 11 covers ``arms in front,'' and pose 7 and 12 are T and A-poses. Note that for the unnatural ``hands down'' poses, the Jacobian and FABRIK choose natural-looking approximations while ONIA achieves best alignment. FABRIK was trapped in a singularity for pose 3. For other poses, all solvers achieve similar alignment, with ONIA being the fastest. See Fig. \ref{fig:poses-metrics} for the metrics evaluated at each pose for every solver. Detailed descriptions and results for each pose and each solver are available on our online appendix: \textcolor{blue}{\protect\url{https://avatar-robots.github.io/poses}}}
    \label{fig:all-poses}
\end{figure}

\section{EVALUATION}

\subsection{Metrics}
\begin{figure*}[t]
    \centering
    \includegraphics[width=\linewidth]{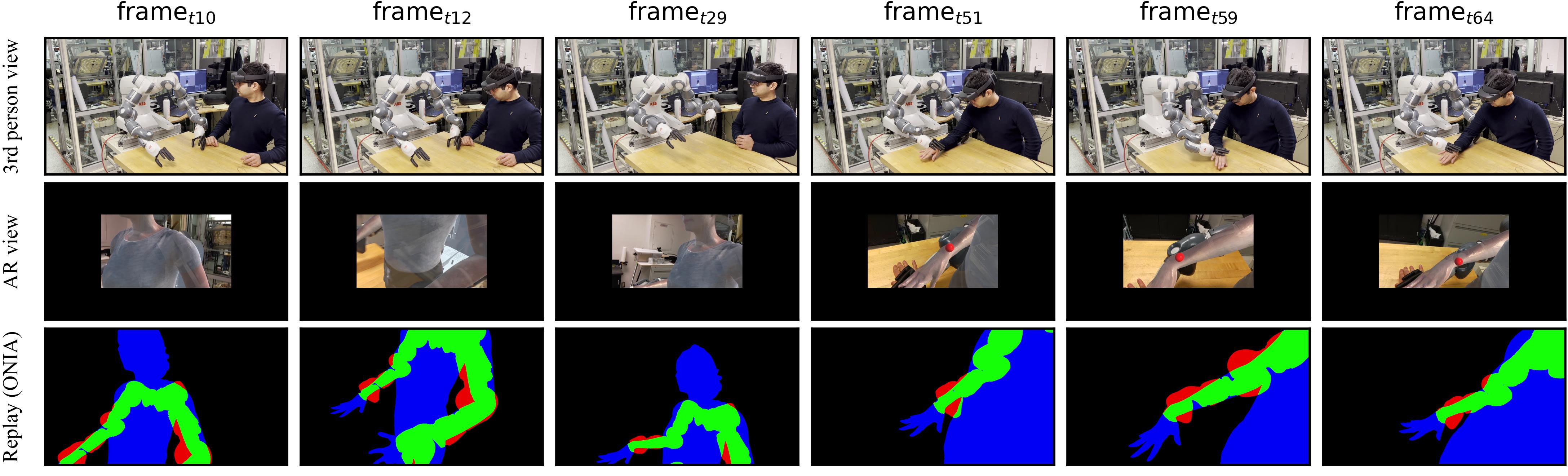}
    \caption{The key frames of the emulated therapy session. The first row shows the session as seen from an external camera; The second row shows the recording of the AR device the user is wearing, which, due to limitation of the recording software, only captures a small central portion of actual user view; The third row shows the replay scene of the session with ONIA, rendered with a special shader, where green represents parts of the robot arm covered by the human model, red represents uncovered area of the robot arm, and blue represents the rest of the human model. The overlay ratio is then the green area over the red and green area combined. See Fig. \ref{fig:therapy-metrics} for the quantitative analysis. The full video of the session with detailed results can be found in our online appendix: \textcolor{blue}{\protect\url{https://avatar-robots.github.io/sessions/emulated-therapy-session}}}
    \label{fig:therapy-key-frames}
\end{figure*}

We define the following metrics in order to evaluate the performance of the solvers in our Avatar Robot System:
\begin{enumerate}
    \item \textbf{Overlay ratio} of a solver at a certain frame is defined as 
    \begin{equation}
        O \coloneq \frac{\text{area of robot arms overlaid by human model}}{\text{total area of robot arms}},
    \end{equation}
    where ``area'' is computed from the patient's perspective, i.e. the area of projection into patient's eye view. This is an intuitive metric of visual alignment: a value of 1 would place all robot arms behind the human model and only the human model would be visible. 
    %This is impossible in our scenario because the avatar mesh is thinner than the robot's arms. 
    We compute the overlay ratio by replaying sessions in simulations with a special shader (see Fig. \ref{fig:therapy-key-frames}).%\footnote{In the session log and the replay, we only know the estimated pose of YuMi and not its actual position. Hence, the computed overlay ratio is a good approximation to how much of robot's arm were covered as seen by the patient.} .
    \item \textbf{Elbow (wrist) deviation}: The elbow deviation $\Delta x_\elbow$ and the wrist deviation $\Delta x_\wrist$ are defined as the distances of the human model's elbow and wrist joints from their corresponding reference positions of the robot arm,
    \begin{align}
        \Delta x_\elbow &\coloneq \norm{\vect x_\shoulder^h + R_\shoulder(S_\upperarm\vect d_\upperarm^h) - \vect x_\elbow^r},\\
        \Delta x_\wrist &\coloneq \norm{\vect x_\shoulder^h + R_\shoulder(S_\upperarm\vect d_\upperarm^h) + R_\elbow(S_\forearm\vect d_\forearm^h) - \vect x_\wrist^r}.
    \end{align}
    
    Smaller deviation should imply better alignment. ONIA is optimal in a sense by ensuring $\Delta x_\elbow = \Delta x_\wrist = 0$, but these metrics still help compare other solvers.
    %
    % While ONIA achieves zero deviation by definition, these metrics still help compare other solvers.
    \item \textbf{Upper \& forearm shrink/stretch}: The scaling factors of the human model's arm segments are $S_\upperarm$ (\ref{eq:upperarm-scale}) and $S_\forearm$ (\ref{eq:forearm-scale}). 
    %E.g., a stretch of 1.2 means the arm segment is scaled to $120\%$ its original length. 
    %Given that the lengths of avatar arm segments do not match those of robot arm linkages, shrinking or stretching is almost always inevitable, but 
    Scaling less aggressively keeps the human model more anthropomorphic. In our plots we present them by their absolute differences to 1 (i.e. $|S_\upperarm-1|$ and $|S_\forearm-1|$), the smaller the difference, the better. 
    \item \textbf{Computation time per frame}: The average time a solver spent on computing alignment pose per frame. Admittedly, the solvers are unlikely to be the performance bottleneck of the entire system since rendering tend to be more resource-intensive. Still, this metric reflects the complexity of different solvers and a faster, simpler one should be preferred if multiple solvers align the human model equally well.
\end{enumerate}

We compute these metrics by logging the robot's position, arm poses, and the patient's perspective (camera transform) of a session and replaying them later in simulation. This replay-based approach allows us to compare multiple solvers controlling for the robot trajectory and patient movement. We evaluate our solvers with these metrics in two ways.

\subsection{Evaluating solvers on a set of static poses}
\label{sec:eval-static}

\begin{figure}[]
    \centering
    \includegraphics[width=\linewidth]{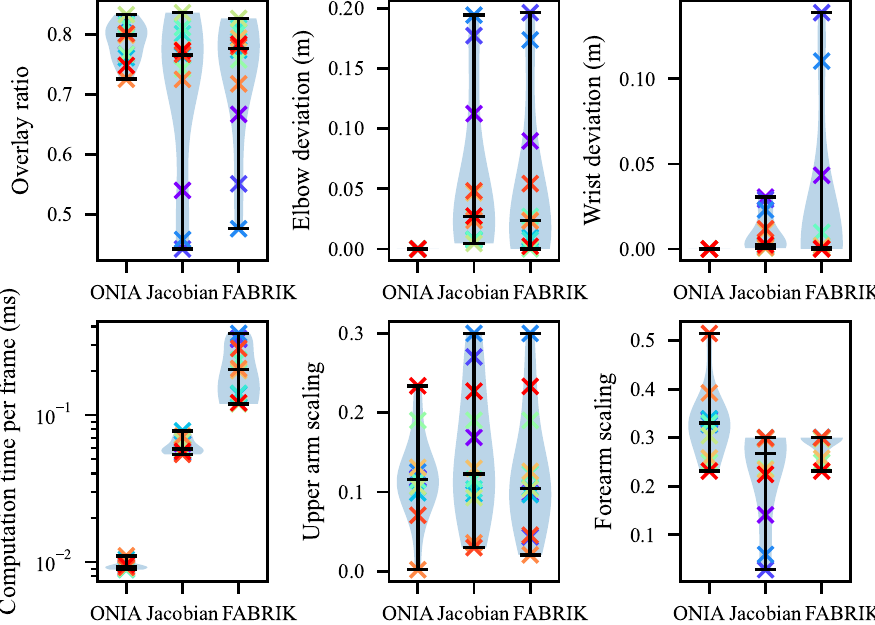}
    \caption{Violin plots of performance metrics of solvers evaluated from 12 poses. The poses are presented in Fig. \ref{fig:all-poses} and marker colors here matches those in Fig. \ref{fig:all-poses}. The minimums, maximums and medians are marked with black bars. We see that ONIA is the fastest of all, achieves the best overlay ratio, and scales the upper arm most conservatively and most consistently.}
    \label{fig:poses-metrics}
\end{figure}

\begin{figure}
    \centering
    \includegraphics[width=\linewidth]{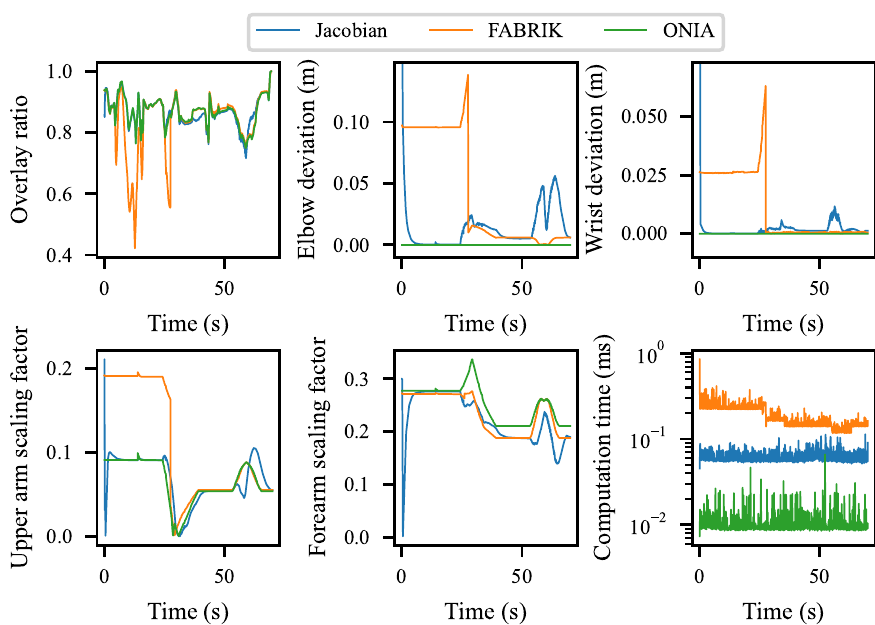}
    \caption{Metrics of all solvers evaluated throughout the therapy session. ONIA has the highest overlay ratio in average and runs the fastest. It scales the upper arm similar to FABRIK but scales the forearm more aggressively, consistent with section \ref{sec:eval-static}. FABRIK exhibits poor alignment metrics for initially, and we find Jacobian to be lagging behind robot movements even with the smallest damping necessary to keep it stable.}
    \label{fig:therapy-metrics}
\end{figure}
We evaluated the performance of the three solvers through a set of 12 static poses, as shown in Fig. \ref{fig:all-poses}. Poses are chosen in group, with each group representative of one partition of the space of all possible robot arm poses. Poses 1 to 3 covers the general case of ``hands down'', and vary in whether the hand points forward, upright, or backward; Likewise, poses 4 to 6 covers the case of ``hands up.''; Poses 8 to 11 covers different cases where the arms are in front of the body; Poses 7 and 12 are the standard T-pose and A-pose.

%\rev{Since our solvers handle each arm separately, we choose the poses to be symmetric so the solvers on both sides are given similar inputs and should produce almost-mirrored poses.} %Taking the mean of the metrics from both sides then effectively reduces the measurement error.

We compute metrics for every pose and each solver. The overlay ratio was evaluated from looking at the robot and human model from the front (as shown in Fig. \ref{fig:all-poses}). We present the results in Fig. \ref{fig:poses-metrics}.

We noticed that even though computational complexity varies greatly, all three solvers achieve similar alignment. Jacobian and FABRIK has similar elbow and wrist deviations. Alignment was poor for both baselines on poses 1 to 3 (the ``hands down'') poses, because these poses are unnatural for human and lie on the boundary of human anatomical constraints. Incorporating these constraints, both baseline solvers produces only natural-looking approximation to the given robot arm pose. Whether a user prefers visually aligned or natural-looking pose is the subject of future user studies. Still, even when \emph{excluding the 3 outliers}, ONIA achieves the highest median overlay ratio (hence best visual alignment) \footnote{Median overlay ratios for 3 solvers excluding poses 1-3 are not shown in Fig. \ref{fig:poses-metrics}. They are .798, .773, and .783 for ONIA, Jacobian and FABRIK respectively.}. Even without explicit scaling constraints, ONIA solver scales the upper arm less aggressively and more consistently (low variance) than both baselines. As of forearm, ONIA tend to scale more aggressively while Jacobian has the most consistent scaling. FABRIK was trapped at a singularity for pose 12, as shown in Fig. \ref{fig:all-poses}, with the arm straight and the back of the elbow facing upwards. In general, we find that FABRIK is more likely to suffer from such singularities than Jacobian. Being non-iterative, ONIA runs the fastest, followed by Jacobian and FABRIK which are each an order of magnitude more computationally expensive then one another.

\subsection{Evaluating the Avatar Robot System sthrough an emulated physical therapy session}
We further evaluated our solvers through an emulated therapy session: `shoulder flexion with elbow extension therapy' \footnote{The full video of the emulated therapy session can be found in our online appendix: \textcolor{blue}{\protect\url{https://avatar-robots.github.io/sessions/emulated-therapy-session/}}} In the emulated therapy, we program the robot to follow a trajectory that guides the user's forearm to rotate around his elbow joint. 
The key frames of the session are shown in Fig. \ref{fig:therapy-key-frames}, and the metrics evaluated at every frame for every solver are presented in Fig. \ref{fig:therapy-metrics}. 

We noticed that ONIA has the best average overlay ratio throughout the session and runs the fastest. It stretches the upper arm of the model similar to FABRIK but stretches the forearm more aggressively. These are all consistent with our previous findings from in section \ref{sec:eval-static}. Jacobian lagged behind changes in robot pose, as shown in its spikes of elbow/wrist deviation when the robot is moving the patient's arm. This is caused by the damping factor of DLS. Reducing the damping further than the current value may improve the responsiveness of Jacobian but causes instability. The damping also causes Jacobian to scale the arm segments differently than the other two solvers did. The left-arm FABRIK solver was caught in a singularity similar to the one it encountered in the static pose evaluation. This caused very poor alignment of the left arm which was reflected in abnormal overlay ratio, elbow/wrist deviations, and upper arm stretching. FABRIK recovered from this singularity on its own in the middle of the session and achieved decent alignment for the second half.

\section{LIMITATIONS}
% We list below the limitations of our approach in this paper, which can provide insights to future investigations:
% \begin{enumerate}
%     \item As outlined in section \ref{sec:preliminary}, all the solvers in our framework consider the alignment of each arm separately. Future alignment solvers could include dual arm poses with the torso to improve the alignment.
%     \item ONIA solver does not limit the scaling of arm segments. While evaluations show that it produces scaling comparable to the baselines in most cases, excessive scaling as high as 1.5 is still observed in pose 11 of Fig. \ref{fig:all-poses}. It would be ideal if a scaling limit could be incorporated into future ONIA versions without significantly compromising its theoretical guarantees.
%     \item For this research, we only considered the alignment of arms between the human model and robot (shoulder, elbow, and wrist joints). The alignment of the hand gesture (including the fingers) to the robot's end effector needs to be explored in the future;
%     \item Our evaluation contains objective metrics only. While this allows precise quantitative analysis of solvers, we expect to benefit from follow-up user studies that can provide new insights about patient perceptions. Subjective evaluations will complement our existing metrics and gives a more comprehensive evaluation of our solvers and our system.
% \end{enumerate}

We list below the limitations of our approach , which can provide insights to future investigations:
\begin{enumerate}
    \item The solvers consider the alignment of each arm separately. Future alignment solvers could include dual arm poses with the torso to improve the alignment.
    \item ONIA solver does not limit the scaling of arm segments. While evaluations show that it produces scaling comparable to the baselines in most cases, excessive scaling as high as 1.5 is still observed in pose 8 of Fig. \ref{fig:all-poses}. It would be ideal if a scaling limit could be incorporated into future ONIA versions without significantly compromising its theoretical guarantees.
    \item For this research, we only considered the alignment of arms between the human model and robot (shoulder, elbow, and wrist joints). The alignment of the hand gesture (including the fingers) to the robot's end effector needs to be explored in the future;
    \item Our evaluation contains objective metrics only. While this allows precise quantitative analysis of solvers, we expect to benefit from follow-up user studies that can provide new insights about patient perceptions. Subjective evaluations will complement our existing metrics and provide a more comprehensive evaluation.
\end{enumerate}

\section{CONCLUSION}
We presented an Avatar Robot System as a mixed real/virtual robotic system that aligns the 3D human model (remote operator) to the robot structure.
Therefore, the care receiver might feel as though they are in close contact with the remote operator.
We proposed a non-iterative alignment solver that optimally aligns arms poses of the human model with robot joints. 
% We gave theoretical guarantees of its quality of alignment and compared it with baseline solvers adapted from iterative Inverse Kinematic solvers (Jacobian DLS and FABRIK). 
We also proposed a set of evaluation metrics that quantitatively measures alignment quality. 
Our evaluation shows that our non-iterative solver consistently achieves alignment quality comparable to the baselines, meanwhile avoiding their limitations such as singularities, instability, and hyper-parameter sensitivity. 


%%%%%%%%%%%%%%%%%%%%%%%%%%%%%%%%%%%%%%%%%%%%%%%%%%%%%%%%%%%%%%%%%%%%%%%%%%%%%%%%

%%%%%%%%%%%%%%%%%%%%%%%%%%%%%%%%%%%%%%%%%%%%%%%%%%%%%%%%%%%%%%%%%%%%%%%%%%%%%%%%

%%%%%%%%%%%%%%%%%%%%%%%%%%%%%%%%%%%%%%%%%%%%%%%%%%%%%%%%%%%%%%%%%%%%%%%%%%%%%%%%
% \section*{Appendix}


%%%%%%%%%%%%%%%%%%%%%%%%%%%%%%%%%%%%%%%%%%%%%%%%%%%%%%%%%%%%%%%%%%%%%%%%%%%%%%%%

\bibliographystyle{IEEEtran}
\bibliography{references}  % .bib

\end{document}

%% file: ref_latex/kinematics.tex
\tikzset{every picture/.style={line width=0.75pt}} %set default line width to 0.75pt        

\begin{tikzpicture}[x=0.75pt,y=0.75pt,yscale=-1,xscale=1]

%Straight Lines [id:da10183285192676972] 
\draw [color={rgb, 255:red, 65; green, 117; blue, 5 }  ,draw opacity=1 ][line width=1.5]  [dash pattern={on 5.63pt off 4.5pt}]  (95.33,79.5) -- (154.4,125.49) ;
\draw [shift={(157.56,127.94)}, rotate = 217.9] [fill={rgb, 255:red, 65; green, 117; blue, 5 }  ,fill opacity=1 ][line width=0.08]  [draw opacity=0] (6.43,-3.09) -- (0,0) -- (6.43,3.09) -- (4.27,0) -- cycle    ;
%Straight Lines [id:da606042482004036] 
\draw [color={rgb, 255:red, 155; green, 155; blue, 155 }  ,draw opacity=1 ] [dash pattern={on 0.84pt off 2.51pt}]  (36,79.17) -- (52.24,79.26) -- (95.33,79.5) ;
%Shape: Arc [id:dp7595240752807368] 
\draw  [draw opacity=0] (207.86,89.25) .. controls (210.36,82.88) and (213.8,78.48) .. (215.87,79.19) .. controls (218.11,79.97) and (217.89,86.52) .. (215.36,93.81) .. controls (212.84,101.1) and (208.97,106.38) .. (206.72,105.6) .. controls (204.95,104.98) and (204.71,100.79) .. (205.93,95.47) -- (211.29,92.4) -- cycle ; \draw [color={rgb, 255:red, 74; green, 144; blue, 226 }  ,draw opacity=1 ]   (209.08,86.44) .. controls (211.4,81.64) and (214.12,78.59) .. (215.87,79.19) .. controls (218.11,79.97) and (217.89,86.52) .. (215.36,93.81) .. controls (212.84,101.1) and (208.97,106.38) .. (206.72,105.6) .. controls (204.95,104.98) and (204.71,100.79) .. (205.93,95.47) ;  \draw [shift={(207.86,89.25)}, rotate = 294.15] [fill={rgb, 255:red, 74; green, 144; blue, 226 }  ,fill opacity=1 ][line width=0.08]  [draw opacity=0] (5.36,-2.57) -- (0,0) -- (5.36,2.57) -- (3.56,0) -- cycle    ;
%Shape: Arc [id:dp8324698365419758] 
\draw  [draw opacity=0] (117.77,79.48) .. controls (118.45,72.4) and (120.67,66.97) .. (122.91,67.12) .. controls (125.28,67.28) and (126.78,73.66) .. (126.26,81.36) .. controls (125.73,89.05) and (123.38,95.16) .. (121.01,95) .. controls (119.14,94.88) and (117.81,90.89) .. (117.59,85.44) -- (121.96,81.06) -- cycle ; \draw [color={rgb, 255:red, 74; green, 144; blue, 226 }  ,draw opacity=1 ]   (118.17,76.42) .. controls (119.11,70.9) and (121,66.99) .. (122.91,67.12) .. controls (125.28,67.28) and (126.78,73.66) .. (126.26,81.36) .. controls (125.73,89.05) and (123.38,95.16) .. (121.01,95) .. controls (119.14,94.88) and (117.81,90.89) .. (117.59,85.44) ;  \draw [shift={(117.77,79.48)}, rotate = 278.26] [fill={rgb, 255:red, 74; green, 144; blue, 226 }  ,fill opacity=1 ][line width=0.08]  [draw opacity=0] (5.36,-2.57) -- (0,0) -- (5.36,2.57) -- (3.56,0) -- cycle    ;
%Shape: Arc [id:dp9468458938835527] 
\draw  [draw opacity=0] (104.62,68.91) .. controls (107.92,71.69) and (109.89,75.9) .. (109.58,80.47) .. controls (109.27,85.01) and (106.78,88.89) .. (103.17,91.2) -- (95.33,79.5) -- cycle ; \draw [color={rgb, 255:red, 74; green, 144; blue, 226 }  ,draw opacity=1 ]   (106.71,71.1) .. controls (108.72,73.69) and (109.82,76.97) .. (109.58,80.47) .. controls (109.34,83.97) and (107.81,87.07) .. (105.47,89.37) ; \draw [shift={(103.17,91.2)}, rotate = 315.76] [fill={rgb, 255:red, 74; green, 144; blue, 226 }  ,fill opacity=1 ][line width=0.08]  [draw opacity=0] (5.36,-2.57) -- (0,0) -- (5.36,2.57) -- (3.56,0) -- cycle    ; \draw [shift={(104.62,68.91)}, rotate = 52.31] [fill={rgb, 255:red, 74; green, 144; blue, 226 }  ,fill opacity=1 ][line width=0.08]  [draw opacity=0] (5.36,-2.57) -- (0,0) -- (5.36,2.57) -- (3.56,0) -- cycle    ;
%Shape: Arc [id:dp5221679920211941] 
\draw  [draw opacity=0] (193.4,83.31) .. controls (199.52,83.64) and (204.18,82.83) .. (204.35,81.26) .. controls (204.56,79.41) and (198.51,77.22) .. (190.84,76.36) .. controls (183.17,75.5) and (176.79,76.29) .. (176.58,78.14) .. controls (176.4,79.74) and (180.93,81.6) .. (187.13,82.62) -- (190.47,79.7) -- cycle ; \draw [color={rgb, 255:red, 74; green, 144; blue, 226 }  ,draw opacity=1 ]   (196.46,83.38) .. controls (201.01,83.33) and (204.21,82.55) .. (204.35,81.26) .. controls (204.56,79.41) and (198.51,77.22) .. (190.84,76.36) .. controls (183.17,75.5) and (176.79,76.29) .. (176.58,78.14) .. controls (176.4,79.74) and (180.93,81.6) .. (187.13,82.62) ;  \draw [shift={(193.4,83.31)}, rotate = 1.85] [fill={rgb, 255:red, 74; green, 144; blue, 226 }  ,fill opacity=1 ][line width=0.08]  [draw opacity=0] (5.36,-2.57) -- (0,0) -- (5.36,2.57) -- (3.56,0) -- cycle    ;
%Straight Lines [id:da010135299455136026] 
\draw [line width=1.5]    (95.33,79.5) -- (185.56,85.35) ;
\draw [shift={(189.56,85.61)}, rotate = 183.71] [fill={rgb, 255:red, 0; green, 0; blue, 0 }  ][line width=0.08]  [draw opacity=0] (6.43,-3.09) -- (0,0) -- (6.43,3.09) -- (4.27,0) -- cycle    ;
%Straight Lines [id:da19625880311948785] 
\draw [line width=1.5]    (189.56,85.61) -- (274.09,113.36) ;
\draw [shift={(277.89,114.61)}, rotate = 198.18] [fill={rgb, 255:red, 0; green, 0; blue, 0 }  ][line width=0.08]  [draw opacity=0] (6.43,-3.09) -- (0,0) -- (6.43,3.09) -- (4.27,0) -- cycle    ;
%Straight Lines [id:da7849065318124158] 
\draw [color={rgb, 255:red, 245; green, 166; blue, 35 }  ,draw opacity=1 ] [dash pattern={on 4.5pt off 4.5pt}]  (191.88,65.42) -- (186.89,108.78) ;
\draw [shift={(192.22,62.44)}, rotate = 96.57] [fill={rgb, 255:red, 245; green, 166; blue, 35 }  ,fill opacity=1 ][line width=0.08]  [draw opacity=0] (5.36,-2.57) -- (0,0) -- (5.36,2.57) -- (3.56,0) -- cycle    ;
%Shape: Polygon [id:ds7101457173934032] 
\draw  [line width=1.5]  (286.56,111.94) -- (277.89,114.61) -- (280.22,124.61) -- (290.89,127.28) -- (295.22,118.61) -- cycle ;
%Straight Lines [id:da5560885616066034] 
\draw [color={rgb, 255:red, 139; green, 87; blue, 42 }  ,draw opacity=1 ][line width=0.75]    (149.88,88.46) -- (181.23,90.42) ;
\draw [shift={(184.22,90.61)}, rotate = 183.58] [fill={rgb, 255:red, 139; green, 87; blue, 42 }  ,fill opacity=1 ][line width=0.08]  [draw opacity=0] (5.36,-2.57) -- (0,0) -- (5.36,2.57) -- (3.56,0) -- cycle    ;
\draw [shift={(146.89,88.28)}, rotate = 3.58] [fill={rgb, 255:red, 139; green, 87; blue, 42 }  ,fill opacity=1 ][line width=0.08]  [draw opacity=0] (5.36,-2.57) -- (0,0) -- (5.36,2.57) -- (3.56,0) -- cycle    ;
%Straight Lines [id:da952225597585431] 
\draw [color={rgb, 255:red, 139; green, 87; blue, 42 }  ,draw opacity=1 ]   (244.4,109.22) -- (271.04,118) ;
\draw [shift={(273.89,118.94)}, rotate = 198.26] [fill={rgb, 255:red, 139; green, 87; blue, 42 }  ,fill opacity=1 ][line width=0.08]  [draw opacity=0] (5.36,-2.57) -- (0,0) -- (5.36,2.57) -- (3.56,0) -- cycle    ;
\draw [shift={(241.56,108.28)}, rotate = 18.26] [fill={rgb, 255:red, 139; green, 87; blue, 42 }  ,fill opacity=1 ][line width=0.08]  [draw opacity=0] (5.36,-2.57) -- (0,0) -- (5.36,2.57) -- (3.56,0) -- cycle    ;
%Straight Lines [id:da3389068913070763] 
\draw [color={rgb, 255:red, 65; green, 117; blue, 5 }  ,draw opacity=1 ][line width=1.5]  [dash pattern={on 5.63pt off 4.5pt}]  (157.56,127.94) -- (97.82,183.55) ;
\draw [shift={(94.89,186.28)}, rotate = 317.05] [fill={rgb, 255:red, 65; green, 117; blue, 5 }  ,fill opacity=1 ][line width=0.08]  [draw opacity=0] (6.43,-3.09) -- (0,0) -- (6.43,3.09) -- (4.27,0) -- cycle    ;
%Straight Lines [id:da837332006845035] 
\draw [color={rgb, 255:red, 65; green, 117; blue, 5 }  ,draw opacity=1 ][line width=1.5]    (91.44,182.22) -- (98.56,190.94) ;
%Straight Lines [id:da2569044931605573] 
\draw [color={rgb, 255:red, 65; green, 117; blue, 5 }  ,draw opacity=1 ][line width=1.5]    (98.56,190.94) -- (92.56,196.28) ;
%Straight Lines [id:da6443491232722434] 
\draw [color={rgb, 255:red, 65; green, 117; blue, 5 }  ,draw opacity=1 ][line width=1.5]    (91.44,182.22) -- (85.56,187.28) ;
%Curve Lines [id:da8452909214014233] 
\draw [color={rgb, 255:red, 128; green, 128; blue, 128 }  ,draw opacity=1 ] [dash pattern={on 4.5pt off 4.5pt}]  (188.89,90.61) .. controls (184.43,107.48) and (179.98,112.5) .. (163.91,124.24) ;
\draw [shift={(161.56,125.94)}, rotate = 324.16] [fill={rgb, 255:red, 128; green, 128; blue, 128 }  ,fill opacity=1 ][line width=0.08]  [draw opacity=0] (5.36,-2.57) -- (0,0) -- (5.36,2.57) -- (3.56,0) -- cycle    ;
%Curve Lines [id:da9150447047438734] 
\draw [color={rgb, 255:red, 128; green, 128; blue, 128 }  ,draw opacity=1 ] [dash pattern={on 4.5pt off 4.5pt}]  (274.89,120.61) .. controls (222.42,159.55) and (186.94,178.89) .. (104.73,186.71) ;
\draw [shift={(102.22,186.94)}, rotate = 354.79] [fill={rgb, 255:red, 128; green, 128; blue, 128 }  ,fill opacity=1 ][line width=0.08]  [draw opacity=0] (5.36,-2.57) -- (0,0) -- (5.36,2.57) -- (3.56,0) -- cycle    ;
%Shape: Circle [id:dp9191822460661072] 
\draw  [fill={rgb, 255:red, 0; green, 0; blue, 0 }  ,fill opacity=1 ] (92.97,79.5) .. controls (92.97,78.2) and (94.03,77.14) .. (95.33,77.14) .. controls (96.64,77.14) and (97.69,78.2) .. (97.69,79.5) .. controls (97.69,80.8) and (96.64,81.86) .. (95.33,81.86) .. controls (94.03,81.86) and (92.97,80.8) .. (92.97,79.5) -- cycle ;
%Shape: Circle [id:dp8138566364969233] 
\draw  [color={rgb, 255:red, 65; green, 117; blue, 5 }  ,draw opacity=1 ][fill={rgb, 255:red, 65; green, 117; blue, 5 }  ,fill opacity=1 ] (77.79,79.38) .. controls (77.79,78.07) and (78.84,77.02) .. (80.15,77.02) .. controls (81.45,77.02) and (82.51,78.07) .. (82.51,79.38) .. controls (82.51,80.68) and (81.45,81.74) .. (80.15,81.74) .. controls (78.84,81.74) and (77.79,80.68) .. (77.79,79.38) -- cycle ;
%Shape: Circle [id:dp4132031218493153] 
\draw  [color={rgb, 255:red, 65; green, 117; blue, 5 }  ,draw opacity=1 ][fill={rgb, 255:red, 65; green, 117; blue, 5 }  ,fill opacity=1 ] (48.88,79.26) .. controls (48.88,77.95) and (49.93,76.9) .. (51.24,76.9) .. controls (52.54,76.9) and (53.6,77.95) .. (53.6,79.26) .. controls (53.6,80.56) and (52.54,81.62) .. (51.24,81.62) .. controls (49.93,81.62) and (48.88,80.56) .. (48.88,79.26) -- cycle ;
%Shape: Circle [id:dp3751280908963679] 
\draw  [fill={rgb, 255:red, 0; green, 0; blue, 0 }  ,fill opacity=1 ] (33.64,79.17) .. controls (33.64,77.86) and (34.7,76.81) .. (36,76.81) .. controls (37.3,76.81) and (38.36,77.86) .. (38.36,79.17) .. controls (38.36,80.47) and (37.3,81.53) .. (36,81.53) .. controls (34.7,81.53) and (33.64,80.47) .. (33.64,79.17) -- cycle ;
%Straight Lines [id:da07261075296230346] 
\draw [color={rgb, 255:red, 155; green, 155; blue, 155 }  ,draw opacity=1 ] [dash pattern={on 0.84pt off 2.51pt}]  (65.33,50.61) -- (65.33,198.28) ;
%Straight Lines [id:da7239501526620495] 
\draw    (262.22,161.28) -- (242.95,124.6) ;
\draw [shift={(241.56,121.94)}, rotate = 62.28] [fill={rgb, 255:red, 0; green, 0; blue, 0 }  ][line width=0.08]  [draw opacity=0] (5.36,-2.57) -- (0,0) -- (5.36,2.57) -- (3.56,0) -- cycle    ;
%Straight Lines [id:da2568364710872255] 
\draw [color={rgb, 255:red, 65; green, 117; blue, 5 }  ,draw opacity=1 ]   (219.22,178.61) -- (156.27,148.89) ;
\draw [shift={(153.56,147.61)}, rotate = 25.27] [fill={rgb, 255:red, 65; green, 117; blue, 5 }  ,fill opacity=1 ][line width=0.08]  [draw opacity=0] (5.36,-2.57) -- (0,0) -- (5.36,2.57) -- (3.56,0) -- cycle    ;

% Text Node
\draw (82.56,58.28) node [anchor=north west][inner sep=0.75pt]    {$\boldsymbol{x}_{s}^{h}$};
% Text Node
\draw (171.19,56.43) node [anchor=north west][inner sep=0.75pt]    {$\boldsymbol{x}_{e}^{h}$};
% Text Node
\draw (273,91.83) node [anchor=north west][inner sep=0.75pt]    {$\boldsymbol{x}_{w}^{h}$};
% Text Node
\draw (103.62,49.19) node [anchor=north west][inner sep=0.75pt]  [font=\fontsize{1em}{1.2em}\selectfont,color={rgb, 255:red, 74; green, 144; blue, 226 }  ,opacity=1 ]  {$R_{s}^{\mathrm{swing}}$};
% Text Node
\draw (129,59.9) node [anchor=north west][inner sep=0.75pt]  [color={rgb, 255:red, 74; green, 144; blue, 226 }  ,opacity=1 ]  {$R_{s}^{\mathrm{twist}}$};
% Text Node
\draw (188.76,47.76) node [anchor=north west][inner sep=0.75pt]  [color={rgb, 255:red, 245; green, 166; blue, 35 }  ,opacity=1 ]  {$\boldsymbol{a}^{h}$};
% Text Node
\draw (201.19,59.43) node [anchor=north west][inner sep=0.75pt]  [color={rgb, 255:red, 74; green, 144; blue, 226 }  ,opacity=1 ]  {$R_{e}^{\mathrm{swing}}$};
% Text Node
\draw (220.86,74.43) node [anchor=north west][inner sep=0.75pt]  [color={rgb, 255:red, 74; green, 144; blue, 226 }  ,opacity=1 ]  {$R_{e}^{\mathrm{twist}}$};
% Text Node
\draw (131.51,85.03) node [anchor=north west][inner sep=0.75pt]  [rotate=-2.67]  {$\boldsymbol{d}_{u}^{h}$};
% Text Node
\draw (219.56,102.94) node [anchor=north west][inner sep=0.75pt]    {$\boldsymbol{d}_{f}^{h}$};
% Text Node
\draw (159.84,92.36) node [anchor=north west][inner sep=0.75pt]  [color={rgb, 255:red, 139; green, 87; blue, 42 }  ,opacity=1 ,rotate=-2.67]  {$S_{u}$};
% Text Node
\draw (249.84,115.36) node [anchor=north west][inner sep=0.75pt]  [color={rgb, 255:red, 139; green, 87; blue, 42 }  ,opacity=1 ,rotate=-2.67]  {$S_{f}$};
% Text Node
\draw (165.56,125.61) node [anchor=north west][inner sep=0.75pt]    {$\textcolor[rgb]{0.25,0.46,0.02}{\boldsymbol{x}_{e}^{r}}$};
% Text Node
\draw (85.89,160.61) node [anchor=north west][inner sep=0.75pt]    {$\textcolor[rgb]{0.25,0.46,0.02}{\boldsymbol{x}_{w}^{r}}$};
% Text Node
\draw (112.51,106.36) node [anchor=north west][inner sep=0.75pt]  [color={rgb, 255:red, 65; green, 117; blue, 5 }  ,opacity=1 ,rotate=-2.67]  {$\boldsymbol{d}_{u}^{r}$};
% Text Node
\draw (130.51,157.36) node [anchor=north west][inner sep=0.75pt]  [color={rgb, 255:red, 65; green, 117; blue, 5 }  ,opacity=1 ,rotate=-2.67]  {$\boldsymbol{d}_{f}^{r}$};
% Text Node
\draw (72.56,83.94) node [anchor=north west][inner sep=0.75pt]    {$\boldsymbol{\textcolor[rgb]{0.25,0.46,0.02}{x}}\textcolor[rgb]{0.25,0.46,0.02}{_{s}^{r}}$};
% Text Node
\draw (225.33,163.89) node [anchor=north west][inner sep=0.75pt]  [color={rgb, 255:red, 0; green, 0; blue, 0 }  ,opacity=1 ] [align=left] {human model arm};
% Text Node
\draw (165.33,181.22) node [anchor=north west][inner sep=0.75pt]  [color={rgb, 255:red, 128; green, 128; blue, 128 }  ,opacity=1 ] [align=left] {\textcolor[rgb]{0.25,0.46,0.02}{reference pose of the robot}};
% Text Node
\draw (61.67,100.72) node [anchor=north west][inner sep=0.75pt]   [align=left] {\textit{shoulder}};
% Text Node
\draw (271.67,132.89) node [anchor=north west][inner sep=0.75pt]   [align=left] {\textit{wrist}};
% Text Node
\draw (188.55,111.07) node [anchor=north west][inner sep=0.75pt]  [rotate=-9.14] [align=left] {\textit{elbow}};
% Text Node
\draw (72.17,39.72) node [anchor=north west][inner sep=0.75pt]   [align=left] {\textit{left side}};
% Text Node
\draw (11.67,39.72) node [anchor=north west][inner sep=0.75pt]   [align=left] {\textit{right side}};

\end{tikzpicture}